\documentclass[10pt]{article}
\DeclareFontShape{OT1}{cmss}{b}{n}{<->ssub*cmss/bx/n}{}
\usepackage{lmodern}
\usepackage{amsmath}
\usepackage[protrusion=true, expansion=false]{microtype}
\usepackage{microtype}
\usepackage{amsfonts}
\usepackage{amssymb}
\usepackage{booktabs}
\usepackage{tabularx}
\usepackage{xcolor}
\usepackage{threeparttable}
\usepackage{graphicx}

\usepackage{booktabs}
\usepackage{array}
\usepackage{ragged2e}

\newcolumntype{C}[1]{>{\centering\arraybackslash}p{#1}}
\usepackage[letterpaper]{geometry}
\usepackage{hicss}
\usepackage{times}
\usepackage[none]{hyphenat}
\usepackage{url}
\usepackage{latexsym}
\usepackage{minted}
\usepackage{indentfirst}
\usepackage{graphicx}
\graphicspath{{images/}}
\usepackage[
    style=apa,
  ]{biblatex}
\title{Document Topic Alignment Metrics for Evaluating Topic Models of Short-Text Public Health Communications on Social Media}

\author{Wangjiaxuan Xin \\
 College of Computing and Informatics\\ 
 The University of North Carolina at Charlotte\\
 {\underline{ wxin@charlotte.edu}} \\ \\
  Yaorong Ge\\
College of Computing and Informatics\\ 
 The University of North Carolina at Charlotte\\
 {\underline{yge@charlotte.edu} } \\ \And
    Shuhua Yin \\
 Department of Public Health and Health Administration\\ 
 The University of North Carolina at Charlotte \\
 {\underline{syin2@charlotte.edu} } \\ \\
 Shi Chen\\
Department of Public Health and Health Administration\\ 
The University of North Carolina at Charlotte \\
 {\underline{schen56@charlotte.edu} } \\}

\date{}

\begin{document}
\maketitle


\begin{abstract}
Topic models are widely used to analyze public health-related social media short texts, yet their evaluation remains dominated by metrics that focus entirely on generated topics alone. There is a lack of metrics that quantitatively assess whether assigned topics meaningfully represent the corresponding short-text posts. We propose Document–Topic Alignment metrics (DoTA), an assignment-aware evaluation framework comprising metrics that measure semantic alignment between documents (posts) and their assigned topics. We also introduce margin-based and discriminative variants that capture topic assignment confidence and distinguishability. We evaluate DoTA across five topic models on three public health--related social media datasets from X and compare DoTA metrics with conventional topic-based metrics. Results show that DoTA provides complementary evaluation cues and aligns meaningfully with human evaluations. These findings establish the need for assignment-aware evaluation and demonstrate that the addition of DoTA enables a more comprehensive and practically meaningful evaluation for assessing short-text topic modeling performance.
\end{abstract}

\subsubsection*{Keywords:}

Social Media Analytics, Topic Modeling, Topic Model Evaluation, Public Health Informatics.

\section{Introduction}
\label{introduction}
Social media platforms have become a major channel for public discourse on health-related issues, especially during health emergencies such as the COVID-19 pandemic \parencite{rauchfleisch2021public, zhang2021understanding}. These platforms capture large-scale, real-time public reactions to official communications, policies, and interventions, offering valuable signals for public health surveillance, risk communication, and policy evaluation \parencite{zhang2021understanding}. However, extracting meaningful insights from such data remains challenging because social media texts are typically short, noisy, informal, and semantically ambiguous.

Topic modeling has been widely adopted to uncover latent thematic structures in large text corpora. Classical probabilistic models such as Latent Dirichlet Allocation (LDA) \parencite{blei2003latent} and Biterm Topic Model (BTM) \parencite{yan2013biterm}, as well as more recent neural and embedding-based approaches such as FASTopic \parencite{wu2024fastopic} and BERTopic \parencite{grootendorst2022bertopic}, have been applied to social media short texts. These methods aim to identify coherent topics that summarize dominant themes in user-generated content, thereby supporting downstream tasks such as discourse trend analysis, public opinion discovery, and sentiment monitoring.

Despite methodological advances in topic modeling, evaluating topic model performance remains an open problem, particularly for short-text social media data. Existing evaluation metrics focus primarily on topic-based properties. Coherence metrics such as $C_v$ \parencite{roder2015exploring} and $C_{\text{UMass}}$ \parencite{mimno2011optimizing} assess the semantic consistency of top keywords within a topic, while Topic Uniqueness ($TU$) \parencite{nan2019topic}, Topic Redundancy ($TR$) \parencite{burkhardt2019decoupling}, and Topic Diversity ($TD$) \parencite{dieng2020topic} measure the distinctiveness and coverage of keywords across topics. Although these metrics provide useful information about topic quality, they do not assess whether a topic meaningfully represents the documents assigned to it.

This limitation is especially important for short-text social media data, where each document (e.g., a post or reply) contains limited contextual information and is therefore more vulnerable to topic misassignment. In such settings, a topic model may generate coherent and well-separated topics while still assigning documents to semantically inappropriate topics. Traditional evaluation metrics may therefore overestimate model quality by overlooking document--topic alignment, i.e., the extent to which a document is semantically consistent with its assigned topic.

Motivated by this gap, we argue that topic model evaluation should be inherently multi-dimensional, encompassing both the quality of the generated topics and the quality of document--topic assignment. To this end, we propose \textit{Document--Topic Alignment} (\textit{DoTA}), an assignment-aware evaluation framework (a set of metrics) that explicitly measures the semantic appropriateness of document-to-topic assignments. \textit{DoTA} quantifies the alignment between a document and its assigned topic using embedding-based similarity in a shared semantic vector space. In addition to a base similarity measure, we introduce margin-based and discriminative variants that account for competing topics, thereby considering assignment ambiguity and distinguishability. Unlike traditional topic-based metrics, \textit{DoTA} operates at the document level and provides a complementary perspective on topic model evaluation based on document–topic alignment. In this study, we applied \textit{DoTA} to public health-related social media data, where the short-text nature of the corpus intensifies assignment challenges, while also examining its potential generalizability across models and datasets.

Guided by this framework, we conducted a comprehensive empirical study on multiple public health-related social media datasets collected from X (formerly Twitter). We compared \textit{DoTA} with conventional topic-based metrics, analyzed their relationships through correlation analysis, and further validated the proposed metrics using human document--topic rating tasks. Our findings show that document-level semantic alignment captures an evaluation dimension that is not adequately reflected by traditional topic-based measures, underscoring the need for a more holistic framework for short-text topic modeling.

Accordingly, this study makes the following contributions:

\begin{itemize}
    \item  We framed the semantic alignment
    of documents and topic assignments as an explicit
    topic-model evaluation dimension.

    \item We introduced and formulated the assignment-aware \textit{DoTA} metric family for evaluating document–topic assignment quality through absolute alignment, competitive margin, and discriminative separation.
    
    \item We conducted extensive experiments across multiple topic models and public health-related short-text datasets, showing that \textit{DoTA} provides complementary evaluation cues beyond traditional topic-based metrics, and further validated these cues through human judgments of document--topic alignment.
\end{itemize}

\section{Related Work}
\label{research_background}

Topic modeling is widely used to uncover latent themes in text, including social media and public health applications \parencite{xin2025improving}. Its evaluation generally relies on \textit{automated metrics} or \textit{human-centered evaluations}. Although human evaluation is often considered the gold standard for interpretability, it is costly and difficult to scale, motivating automatic alternatives \parencite{chang2009reading, newman2010automatic, lau2014machine, zhang2026lira}.

Early likelihood-based measures such as perplexity often poorly reflect human interpretability \parencite{chang2009reading}, leading to coherence metrics such as PMI, NPMI, $C_v$ \parencite{roder2015exploring}, and $C_{\text{UMass}}$ \parencite{mimno2011optimizing}. These metrics assess semantic relatedness among topic keywords but may not align with human judgments, particularly for neural topic models \parencite{doogan2021topic, hoyle2021automated}. More importantly, evaluating topics alone overlooks document--topic assignments, even though topic models produce both topic representations and document assignments \parencite{airoldi2016improving, hoyle2021automated}. Thus, coherent topics may still poorly represent their assigned documents.

Recent approaches broaden evaluation toward practical document-level relevance. PROXANN integrates topic identification with document relevance assessment \parencite{hoyle2025proxann}, while WALM uses Large Language Model (LLM)-generated summaries to jointly assess topic and document quality \parencite{yang2025llm}. However, these approaches rely on proxy annotations or LLM-generated content rather than directly quantifying fine-grained document--topic alignment. This motivates an explicit and scalable assignment-aware evaluation framework.

To address these challenges, \textit{DoTA} directly evaluates document–topic semantic alignment. While embedding-based similarity itself is well established, our contribution is to formulate assignment alignment as an explicit topic-model evaluation dimension and characterize it through absolute, margin-based, and discriminative measures.

\section{Methodology - \textit{DoTA} Metrics}
\label{methodology}

\subsection{Problem Setup and Notation}

\begin{figure}[t]
    \centering
    \includegraphics[width=\linewidth]{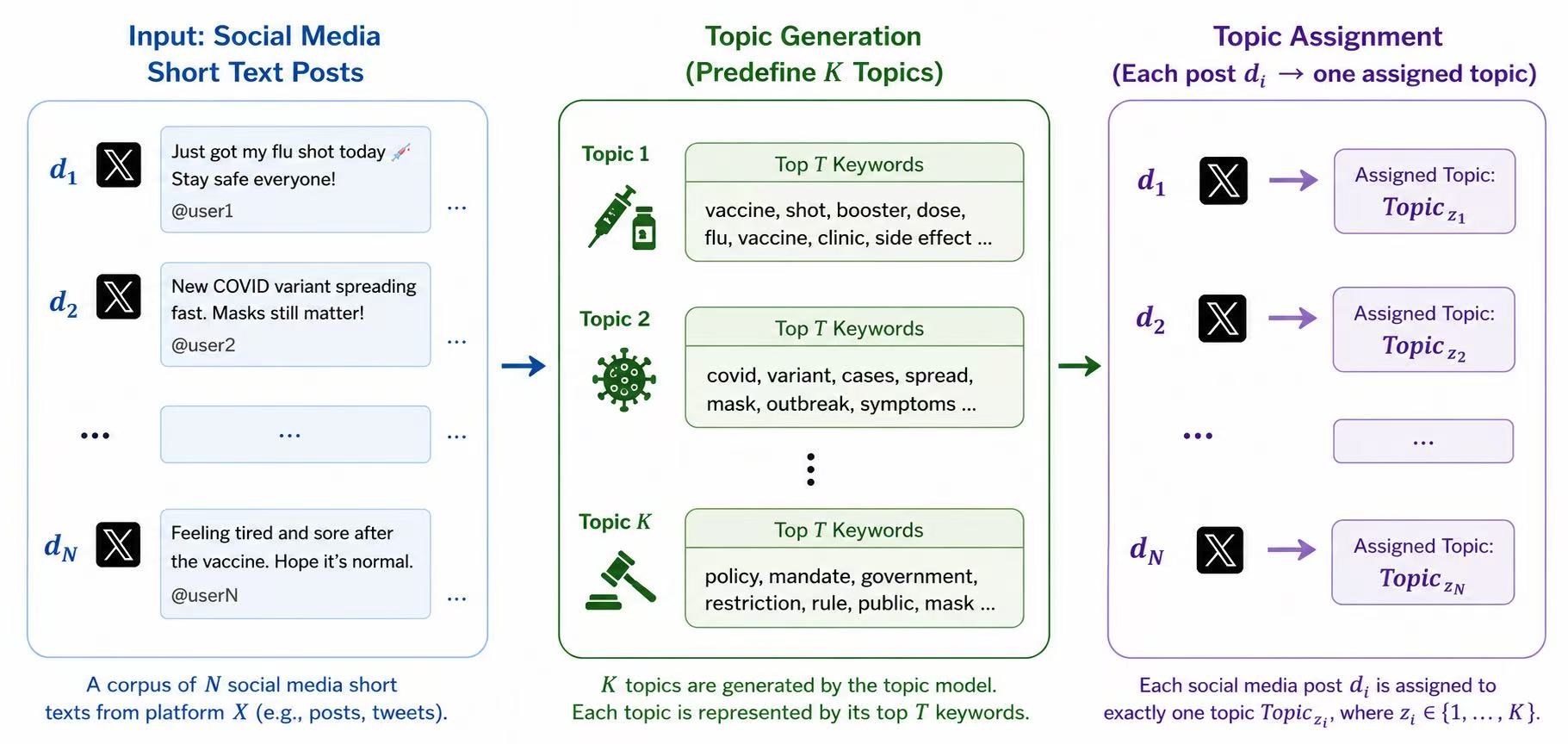}
    \caption{Topic modeling for social media short texts. A corpus of $N$ short posts is used to generate $K$ topics, each represented by $T$ keywords. Each post is then assigned to exactly one topic.}
    \label{fig:topic_modeling}
\end{figure}

Let $\mathcal{D} = \{d_1, d_2, \dots, d_N\}$ denote a corpus of $N$ social media short-text posts (e.g., posts from X). As illustrated in Figure~\ref{fig:topic_modeling}, the topic modeling process in short-text settings can be described in three stages: (i) input corpus, (ii) topic generation, and (iii) topic assignment.

First, given the corpus $\mathcal{D}$, a topic model learns a set of $K$ latent topics $\mathcal{T} = \{t_1, t_2, \dots, t_K\}$. Each topic $t_k$ is represented by a set of top $T$ keywords:
\begin{equation}
t_k = \{w_{1}^{k}, w_{2}^{k}, \dots, w_{T}^{k}\}
\end{equation}

These keywords provide a compact semantic description of the topic.

In contrast to long-document topic modeling, where each document is modeled as a distribution over multiple topics, short-text topic modeling typically assumes a single-topic assignment due to limited contextual information in each post. Specifically, each document (i.e., a post or a reply) $d_i$ is assigned to exactly one topic:
\begin{equation}
z_i \in \{1, \dots, K\}
\end{equation}

where $z_i$ denotes the index of the topic assigned to document $d_i$. To evaluate semantic alignment between documents and topics, we define a common external evaluation space using a pretrained sentence encoder $f(\cdot)$, i.e., \texttt{all-MiniLM-L6-v2} \parencite{reimers2019sentence} in this study, which maps both documents and topics into $\mathbb{R}^D$:
\begin{equation}
\mathbf{d}_i = f(d_i), \quad \mathbf{t}_k = f(\text{concat}(t_k))
\end{equation}

where $\text{concat}(t_k)$ denotes the concatenation of the $T$ keywords of topic $t_k$ into a single sequence, and $D$ is the embedding dimension (i.e., $D=384$ for \texttt{all-MiniLM-L6-v2}).

The semantic similarity between a document and a topic is measured by their cosine similarity:
\begin{equation}
\text{sim}(\mathbf{d}_i, \mathbf{t}_k) = 
\frac{\mathbf{d}_i \cdot \mathbf{t}_k}
{\|\mathbf{d}_i\| \, \|\mathbf{t}_k\|}
\end{equation}

Based on this formulation, we define \textit{DoTA} metrics to quantify the semantic alignment between each short-text document and its assigned topic and non-assigned topics.

\subsection{$DoTA_{\text{base}}$}

The basic document-level \textit{DoTA} metric measures the absolute semantic alignment between a document and its assigned topic by their cosine similarity, i.e., $DoTA_{\text{base}}(d_i) = \text{sim}(\mathbf{d}_i, \mathbf{t}_{z_i})$.

The corpus-level \textit{DoTA} metric is computed as the average of all document-level \textit{DoTA} scores over all documents:
\begin{equation}
DoTA_{\text{base}} = \frac{1}{N} \sum_{i=1}^{N} DoTA_{\text{base}}(d_i)
\end{equation}

This metric captures the extent to which the assigned topic semantically represents the document in the shared embedding space. However, it does not consider alternative topics. Because each post is assigned to exactly one topic due to its short-text nature, other alternative topics may also exhibit high similarity to the document, leading to potential overestimation of assignment quality in ambiguous cases.

\subsection{$DoTA_{\text{max}}$ (Margin-Based Version)}

To account for the most competitive alternative topic, we next introduce a margin-based variant:
\begin{equation}
DoTA_{\text{max}}(d_i) =
\text{sim}(\mathbf{d}_i, \mathbf{t}_{z_i})
- \max_{k \neq z_i} \text{sim}(\mathbf{d}_i, \mathbf{t}_k)
\end{equation}

The corpus-level score is:
\begin{equation}
DoTA_{\text{max}} = \frac{1}{N} \sum_{i=1}^{N} DoTA_{\text{max}}(d_i)
\end{equation}

This metric captures the confidence margin between the assigned topic and the strongest competing alternative (i.e., the topic $t_k$ with the highest cosine similarity with the document $d_i$ among the $K-1$ non-assigned topics), reflecting assignment ambiguity in the hardest case. A higher value indicates clearer and more confident assignments, whereas small or negative margins suggest potential misalignment or confusion between closely related topics. Because $DoTA_{max}$ is an external evaluation criterion rather than the topic model’s assignment objective, a negative value reflects disagreement between the model assignment and the nearest topic in the \textit{DoTA} semantic space.

\subsection{$DoTA_{\text{disc}}$ (Discriminative Version)}

Beyond the single strongest competitor, we further introduce a discriminative variant that evaluates the assigned topic against all alternative topics:
\begin{equation}
DoTA_{\text{disc}}(d_i) = 
\text{sim}(\mathbf{d}_i, \mathbf{t}_{z_i}) 
- \frac{1}{K-1} \sum_{k \neq z_i} \text{sim}(\mathbf{d}_i, \mathbf{t}_k)
\end{equation}

Here, the non-assigned topics serve as competing reference topics for evaluating how strongly the assigned topic stands out on average. The corpus-level score is:
\begin{equation}
DoTA_{\text{disc}} = \frac{1}{N} \sum_{i=1}^{N} DoTA_{\text{disc}}(d_i)
\end{equation}

This formulation measures the discriminative separation of the assigned topic from all alternative topics, providing a broader indication of assignment quality than absolute similarity alone. A higher value suggests that the assigned topic is more distinguishable from the overall topic space, while a lower value indicates greater ambiguity.

\subsection{Design Rationale}

The proposed \textit{DoTA} metrics capture document--topic alignment from a perspective that is complementary to commonly used topic-based evaluation metrics. Rather than relying on a single notion of semantic similarity, \textit{DoTA} is designed as a multi-dimensional evaluation framework with progressively stricter criteria of topic assignment quality. Table~\ref{tab:dota_rationale} summarizes the three \textit{DoTA} variants in terms of evaluation focus, degree of topic competition considered, and interpretation. Together, they form a hierarchy from absolute semantic fit, to margin-based comparison with the strongest competing topic, to broader discriminative separation from all alternative topics.

\begin{table}[t]
\centering
\small
\caption{Comparison of DoTA variants}
\label{tab:dota_rationale}

\setlength{\tabcolsep}{3pt}
\renewcommand{\arraystretch}{1.12}

\begin{tabular}{
@{}
>{\RaggedRight\arraybackslash}p{0.18\columnwidth}
>{\RaggedRight\arraybackslash}p{0.24\columnwidth}
>{\RaggedRight\arraybackslash}p{0.22\columnwidth}
>{\RaggedRight\arraybackslash}p{0.3\columnwidth}
@{}
}
\toprule
\textbf{Metric} & \textbf{Focus} & \textbf{Competition} & \textbf{Interpretation} \\
\midrule
$DoTA_{\mathrm{base}}$
& Absolute Alignment
& None
& Semantic Fit \\

$DoTA_{\mathrm{max}}$
& Confidence Margin
& Hardest Topic
& Ambiguity \\

$DoTA_{\mathrm{disc}}$
& Discriminative Separation
& All Topics
& Distinguish\-ability \\
\bottomrule
\end{tabular}
\end{table}

In summary, $DoTA_{\text{base}}$ measures the direct semantic similarity between a document and its assigned topic, providing a basic estimate of semantic fit. Because it does not consider competing alternative topics, it may overestimate assignment quality in ambiguous cases. $DoTA_{\text{max}}$ focuses on the hardest competing topic, quantifying the confidence margin between the assigned topic and its closest alternative. It is therefore more sensitive to ambiguous or borderline assignments. $DoTA_{\text{disc}}$ incorporates all competing topics by comparing the assigned topic with the average similarity of non-assigned topics. It captures the preference of the assignment and is more robust when topics overlap semantically.

\begin{figure}
    \centering
    \includegraphics[width=1\linewidth]{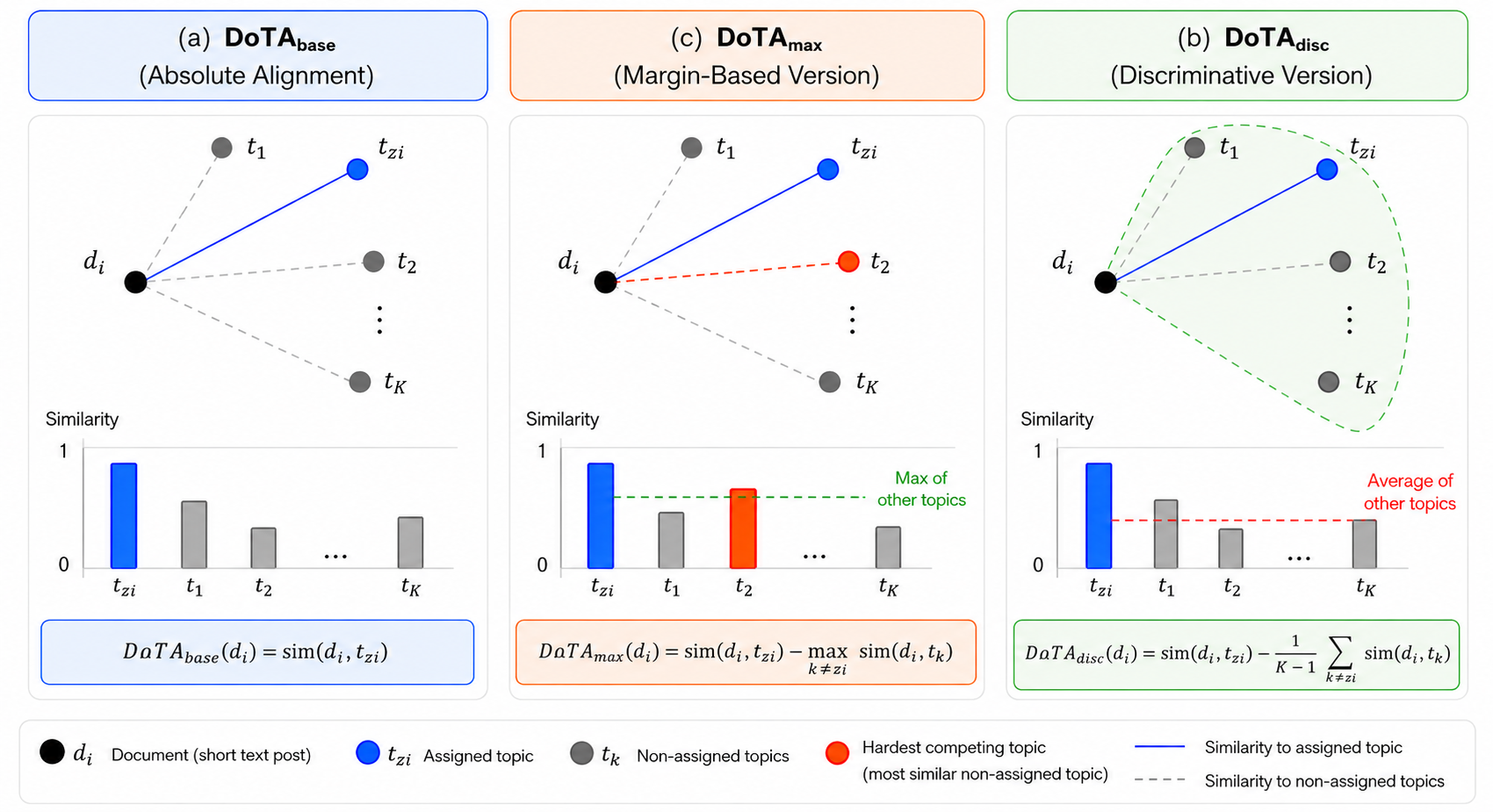}
    \caption{Illustration of three DoTA metrics for a document $d_i$ with assigned topic $t_{z_i}$. (a) $DoTA_{\text{base}}$ measures absolute similarity to the assigned topic. (b) $DoTA_{\text{max}}$ considers the margin between the assigned topic and the most similar competing topic. (c) $DoTA_{\text{disc}}$ compares the similarity to the assigned topic with the average similarity to all other topics.}
    \label{dota_metrics}
\end{figure}

As illustrated in Figure~\ref{dota_metrics}, all three \textit{DoTA} metrics should be interpreted as embedding-space measures of document–topic semantic alignment rather than ground-truth assignment correctness. Specifically, $DoTA_{\text{base}}$, $DoTA_{\text{max}}$, and $DoTA_{\text{disc}}$ respectively capture absolute semantic fit, relative confidence against the strongest competitor, and discriminative separation from alternative topics. These signals complement, rather than replace, topic coherence, interpretability, and broader model-selection criteria.

\section{Experimental Setup}
\label{experiments}

\subsection{Datasets} \label{dataset}
\begin{itemize}
    \item \textbf{Highly Engaged COVID-19 Discussion:} 
    A dataset of highly engaged COVID-19 discussions on X during January 2020 - May 2023 was collected via Brandwatch\footnote{\url{https://www.brandwatch.com/}}. Posts with an engagement score $\geq$ 10 (sum of likes, reposts, and replies) were retained to reflect highly engaged discussions. The final dataset contains 67,895 posts.

    \item \textbf{DoxyPEP Discussion:} 
    An English-language dataset of discussions on doxycycline and its use as post-exposure prophylaxis (DoxyPEP) for sexually transmitted infections (STIs) was also collected from X via Brandwatch. The dataset contains 86,829 U.S. English-language posts published from January 2022 to October 2025 and was retrieved via Brandwatch in December 2025 using queries covering \texttt{doxycycline}, \texttt{doxy-PEP}, and \texttt{STI}-related terms.
    

    \item \textbf{Mpox Discussion:} 
   We also retrieved a publicly available dataset of Monkeypox-related tweets\footnote{\url{https://www.kaggle.com/datasets/vencerlanz09/monkeypox-tweets}} collected between August and September 2022. Of the 93,968 tweets collected, 82,824 English posts were retained, providing a focused snapshot of public discourse during the Monkeypox outbreak.
\end{itemize}

To comply with the X platform’s privacy policy, data collection was limited to publicly accessible posts only. Therefore, approval from the authors’ Institutional Review Board was not required for this study.



\subsection{Topic Models}
\label{topic_models}

To evaluate the proposed \textit{DoTA} metrics against more conventional topic-based metrics across diverse modeling paradigms, we experimented with five representative topic models: LDA, BTM, FASTopic, BERTopic, and topic-semantic contrast topic model (TSCTM), spanning probabilistic and embedding-based neural frameworks. These topic models enable a comprehensive assessment of document-topic alignment under different modeling assumptions.
\begin{itemize}
    \item \textbf{LDA} is a classical probabilistic model that represents documents as mixtures of topics, and topics as word distributions \parencite{blei2003latent}. Despite its assumptions for longer texts, it remains a widely used and interpretable baseline.

    \item \textbf{BTM} is designed for short texts by modeling global word co-occurrence patterns (biterms), effectively alleviating data sparsity issues \parencite{yan2013biterm}.

    \item \textbf{FASTopic} \parencite{wu2024fastopic} is a neural network approach that leverages embeddings from pre-trained models and directly captures semantic relationships between document embeddings and learnable topic–word embeddings to generate coherent topics, offering scalability and robustness for short-text data.

    \item \textbf{BERTopic} combines sentence-transformer embeddings, dimension reduction, and the clustering algorithm, with topic representations derived via class-based TF-IDF (topic frequency-inverse document frequency), providing excellent topic modeling performance \parencite{grootendorst2022bertopic}.

    \item \textbf{TSCTM} employs contrastive learning to address short-text sparsity, producing semantically rich topic representations with improved coherence \parencite{wu2022mitigating}.
\end{itemize}

Each document used the topic assignment produced by its corresponding model, selecting the highest-probability/score topic when multiple topic scores were available; otherwise, the model-provided topic label was used. All models used the same preprocessed corpora and topic-number settings. BTM was implemented with bitermplus\footnote{\url{https://github.com/maximtrp/bitermplus}}, and the other models with TopMost\footnote{\url{https://github.com/bobxwu/TopMost}}.

\subsection{Topic-Based Quantitative Metrics}
\label{topic_metrics}

To complement the proposed \textit{DoTA} metrics, five widely used topic-based quantitative metrics were employed for correlation analysis, including topic coherence ($C_v$ and $C_{\text{UMass}}$), topic uniqueness ($TU$), topic redundancy ($TR$), and topic diversity ($TD$). These metrics evaluate topic quality from different perspectives, such as semantic coherence and lexical distinctiveness.

\begin{itemize}

\item \textbf{$C_v$} \parencite{roder2015exploring}: 
evaluates semantic coherence among topic keywords using NPMI. For a word pair $(w_i, w_j)$:
\begin{equation}
\text{NPMI}(w_i, w_j) =
\frac{\log \left(\frac{P(w_i, w_j) + \epsilon}{P(w_i)P(w_j) + \epsilon}\right)}
{-\log (P(w_i, w_j) + \epsilon)}
\end{equation}

where $\epsilon$ (e.g., $10^{-12}$) avoids numerical instability. The $C_v$ score is:
\begin{equation}
C_V = \frac{1}{T} \sum_{i=1}^{T} 
\cos \big( \mathbf{v}_{\text{NPMI}}(x_i), 
\mathbf{v}_{\text{NPMI}}(\{x_j\}_{j=1}^{T}) \big)
\end{equation}

where
    \begin{equation}
        \mathbf{v}_{\text{NPMI}}(x_i) = 
        \big\{ \text{NPMI}(x_i, x_j) \big\}_{j=1,\dots,T}
    \end{equation}

and
    \begin{equation}
        \mathbf{v}_{\text{NPMI}}(\{x_j\}_{j=1}^{T}) =
        \left\{ \sum_{i=1}^{T} \text{NPMI}(x_i, x_j) \right\}_{j=1,\dots,T}
    \end{equation}

Higher $C_v$ indicates stronger semantic coherence.

\item \textbf{$C_{\text{UMass}}$} \parencite{mimno2011optimizing}: measures coherence via document co-occurrence:
\begin{equation}
C_{\text{UMass}} = \frac{2}{T(T-1)} \sum_{i<j}
\log \frac{D(w_i, w_j) + \epsilon}{D(w_j)}
\end{equation}

where $D(w_i, w_j)$ is the co-occurrence count and $D(w_j)$ is the document frequency. Higher (less negative) values indicate better coherence.

\item \textbf{$TU$} \parencite{nan2019topic}: quantifies how distinct topic keywords are:
\begin{equation}
TU = \frac{1}{K} \sum_{k=1}^{K} 
\left( \frac{1}{T} \sum_{x_i \in T_k} \frac{1}{\#(x_i)} \right)
\end{equation}

where $T_k$ denotes the top keyword set of the $k$-th topic $t_k$ and $\#(x_i)$ is the frequency of $x_i$ across all topic keyword sets. Higher $TU$ indicates greater uniqueness and clearer separation of topics.

\item \textbf{$TR$} \parencite{burkhardt2019decoupling}: measures keyword overlap:
\begin{equation}
TR = \frac{1}{K} \sum_{k=1}^{K} 
\left( \frac{1}{T} \sum_{x_i \in T_k} \frac{\#(x_i) - 1}{K - 1} \right)
\end{equation}

Lower $TR$ implies less redundancy.

\item \textbf{$TD$} \parencite{dieng2020topic}: measures the proportion of unique keywords:
\begin{equation}
TD = \frac{1}{K} \sum_{k=1}^{K} 
\frac{1}{T} \sum_{x_i \in T_k} \mathbb{I}(\#(x_i))
\end{equation}

where
\begin{equation}
\mathbb{I}(\#(x_i)) =
\begin{cases}
1, & \text{if } \#(x_i) = 1 \\
0, & \text{otherwise}
\end{cases}
\end{equation}

Higher $TD$ indicates more diverse and less overlapping topics.

\end{itemize}

These metrics primarily evaluate the performance of topic models at the topic level, whereas \textit{DoTA} metrics focus on document-topic semantic alignment. Such joint analysis enables a comprehensive understanding of topic model performance from both topic-centric and assignment-aware perspectives. 


In the main experiments, we set the number of topics to $K \in \{5,10\}$ to capture two practical levels of topic granularity for short-text public health discussions: a coarser setting and a moderately finer setting. These values were selected for consistent cross-model comparison and to reduce the risk of over-fragmented topics in short-text corpora. For each topic, the top 15 keywords ($T=15$) were retained for metric calculation to ensure a fixed and comparable topic representation across models.

\section{Results and Discussion}

\subsection{Topic-based Metrics and \textit{DoTA} Metrics}

\begin{table*}[t]
\centering
\small

\begin{threeparttable}
\caption{Topic-Based metrics across datasets and topic numbers\tnote{A, B}}
\label{tab:topic_metrics}

\setlength{\tabcolsep}{4pt}  
\begin{tabular}{l|ccccc|ccccc|ccccc}
\toprule
 & \multicolumn{5}{c|}{COVID-19 ($K=5$)} 
 & \multicolumn{5}{c|}{DoxyPEP ($K=5$)} 
 & \multicolumn{5}{c}{Mpox ($K=5$)} \\
Model & $C_v$ & $C_{\text{UMass}}$ & $TU$ & $TR$ & $TD$ 
      & $C_v$ & $C_{\text{UMass}}$ & $TU$ & $TR$ & $TD$
      & $C_v$ & $C_{\text{UMass}}$ & $TU$ & $TR$ & $TD$ \\
\midrule
FASTopic & \textbf{.403} & -6.245 & \textbf{1} & \textbf{0} & \textbf{1} & .233 & -7.146 & \textbf{1} & \textbf{0} & \underline{.96} & .282 & -5.784 & \textbf{1} & \textbf{0} & \textbf{1} \\
BERTopic & \underline{.395} & \underline{-1.803} & .6 & .386 & .6 & .228 & -5.447 & .693 & .24 & .693 & .358 & \textbf{-2.162} & .573 & .38 & .573 \\
TSCTM    & .301 & -4.597 & \textbf{1} & \textbf{0} & \textbf{1} & \underline{.279} & -6.584 & \textbf{1} & \textbf{0} & \textbf{1} & .293 & -5.524 & \textbf{1} & \textbf{0} & \textbf{1} \\
LDA      & .357 & -2.604 & \underline{.666} & \underline{.273} & \underline{.666} & \textbf{.317} & \textbf{-3.797} & .906 & \underline{.046} & .906 & \textbf{.399} & \underline{-2.429} & .72 & .18 & .72 \\
BTM      & .373 & \textbf{-1.739} & .613 & .306 & .613 & .258 & \underline{-4.259} & \underline{.92} & \underline{.046} & .92 & \underline{.377} & -3.7 & \underline{.773} & \underline{.173} & \underline{.773} \\
\midrule
 & \multicolumn{5}{c|}{COVID-19 ($K=10$)} 
 & \multicolumn{5}{c|}{DoxyPEP ($K=10$)} 
 & \multicolumn{5}{c}{Mpox ($K=10$)} \\
 
Model & $C_v$ & $C_{\text{UMass}}$ & $TU$ & $TR$ & $TD$ 
      & $C_v$ & $C_{\text{UMass}}$ & $TU$ & $TR$ & $TD$
      & $C_v$ & $C_{\text{UMass}}$ & $TU$ & $TR$ & $TD$ \\
\midrule
FASTopic & .318 & -6.374 & \textbf{1} & \textbf{0} & \textbf{1} & .269 & -7.625 & \underline{.993} & \underline{.001} & \underline{.986} & .328 & -5.548 & \textbf{.993} & \textbf{.001} & \textbf{.993} \\
BERTopic & .369 & -2.408 & \underline{.593} & \underline{.247} & \underline{.593} & \underline{.273} & -5.504 & .64 & .18 & .64 & .335 & \underline{-2.851} & .586 & .235 & .586 \\
TSCTM    & .341 & -4.689 & \textbf{1} & \textbf{0} & \textbf{1} & .249 & -6.019 & \textbf{1} & \textbf{0} & \textbf{.993} & .283 & -5.081 & \underline{.986} & \underline{.003} & \underline{.986} \\
LDA      & \textbf{.377} & \underline{-2.188} & .566 & .253 & .566 & \textbf{.283} & \underline{-5.007} & .793 & .06 & .793 & \underline{.362} & \textbf{-2.64} & .686 & .115 & .68 \\
BTM      & \underline{.37} & \textbf{-1.538} & .533 & .281 & .533 & .271 & \textbf{-3.865} & .906 & .025 & .9 & \textbf{.377} & -3.308 & .62 & .191 & .62 \\
\bottomrule
\end{tabular}

\begin{tablenotes}[flushleft]
\footnotesize
\item[A] The best performance within each dataset-$K$ pair is in \textbf{bold} and the second best is \underline{highlighted}. 
\item[B] Results are averaged on five independent runs. 
\end{tablenotes}

\end{threeparttable}
\end{table*}

\begin{table*}[t]
\centering
\small

\begin{threeparttable}
\caption{DoTA metrics across datasets and topic numbers\tnote{A, B}}
\label{tab:dota_metrics}

\setlength{\tabcolsep}{3pt}
\begin{tabular}{l|ccc|ccc|ccc}
\toprule
 & \multicolumn{3}{c|}{COVID-19 ($K=5$)} 
 & \multicolumn{3}{c|}{DoxyPEP ($K=5$)} 
 & \multicolumn{3}{c}{Mpox ($K=5$)} \\
Model & $DoTA_{\text{base}}$ & $DoTA_{\text{max}}$ & $DoTA_{\text{disc}}$ 
      & $DoTA_{\text{base}}$ & $DoTA_{\text{max}}$ & $DoTA_{\text{disc}}$ 
      & $DoTA_{\text{base}}$ & $DoTA_{\text{max}}$ & $DoTA_{\text{disc}}$ \\
\midrule
FASTopic & .17 & -.054 & \underline{.074} & .292 & \underline{.052} & .152 & .206 & \underline{.001} & \textbf{.099} \\
BERTopic & \textbf{.385} & \textbf{-.017} & .06 & \textbf{.466} & .014 & .103 & \textbf{.54} & \textbf{.043} & \textbf{.099} \\
TSCTM    & .21 & -.064 & \textbf{.076} & .328 & .037 & .157 & .215 & -.018 & .077 \\
LDA      & \underline{.365} & \underline{-.035} & .029 & \underline{.464} & \textbf{.126} & \textbf{.229} & .381 & -.065 & \underline{.098} \\
BTM      & .34 & -.057 & .017 & .433 & .015 & \underline{.191} & \underline{.469} & -.037 & .032 \\
\midrule
 & \multicolumn{3}{c|}{COVID-19 ($K=10$)} 
 & \multicolumn{3}{c|}{DoxyPEP ($K=10$)} 
 & \multicolumn{3}{c}{Mpox ($K=10$)} \\
Model & $DoTA_{\text{base}}$ & $DoTA_{\text{max}}$ & $DoTA_{\text{disc}}$ 
      & $DoTA_{\text{base}}$ & $DoTA_{\text{max}}$ & $DoTA_{\text{disc}}$ 
      & $DoTA_{\text{base}}$ & $DoTA_{\text{max}}$ & $DoTA_{\text{disc}}$ \\
\midrule
FASTopic & .203 & -.059 & \textbf{.111} & .331 & -.026 & .187 & .248 & \underline{-.014} & .104 \\
BERTopic & \textbf{.386} & \textbf{-.025} & .076 & \textbf{.493} & -.024 & .155 & \textbf{.527} & \textbf{.007} & \textbf{.123} \\
TSCTM    & .265 & -.142 & \underline{.078} & .306 & \underline{.007} & .181 & .263 & -.027 & .113 \\
LDA      & \underline{.366} & \underline{-.056} & .032 & \underline{.465} & \textbf{.014} & \underline{.195} & .359 & -.076 & \underline{.119} \\
BTM      & .324 & -.101 & .021 & .434 & -.051 & \textbf{.215} & \underline{.471} & -.066 & .066 \\
\bottomrule
\end{tabular}

\begin{tablenotes}[flushleft]
\footnotesize
\item[A] The best performance within each dataset-$K$ pair is in \textbf{bold} and the second best is \underline{highlighted}. 
\item[B] Results are averaged on five independent runs. 
\end{tablenotes}

\end{threeparttable}
\end{table*}

The results in Tables~\ref{tab:topic_metrics} and \ref{tab:dota_metrics} reveal several important limitations of conventional topic-based evaluation metrics when applied to short-text social media data. First, metrics such as $TU$, $TR$, and $TD$ exhibit similar patterns and high correlation across models and datasets. For instance, both FASTopic and TSCTM consistently achieve extreme values (e.g., $TU=1$, $TR=0$, $TD=1$), suggesting near-perfect topic distinctiveness. However, these uniformly high scores fail to meaningfully differentiate model performance. This indicates that such structural metrics are often insensitive to nuanced variations in topic quality and may overestimate performance by rewarding lexical separation rather than semantic relevance.

In addition, coherence metrics ($C_v$ and $C_{\text{UMass}}$) demonstrate inconsistent behavior across datasets and topic settings. While topic models (e.g., FASTopic) achieve competitive coherence scores, these improvements do not consistently translate into better semantic alignment at the document level. For example, models with strong $C_v$ scores do not necessarily produce stronger document–topic semantic alignment, highlighting a discrepancy between topic-based interpretability and document–topic semantic alignment. This limitation is particularly pronounced in short-text settings, where sparse context reduces the reliability of word co-occurrence signals.

In contrast, Table~\ref{tab:dota_metrics} shows that the proposed \textit{DoTA} metrics provide more discriminative and informative evaluation signals. Unlike conventional metrics, \textit{DoTA} explicitly measures document--topic semantic alignment. For example, in the COVID-19 dataset with $K=5$, $DoTA_{\text{base}}$ clearly distinguishes BERTopic (.385) and LDA (.365) from FASTopic (.17) and TSCTM (.21), even though the latter achieve near-perfect scores on structural metrics such as $TU$ and $TD$. A similar pattern appears across datasets: for DoxyPEP ($K=5$), BERTopic (.466) and LDA (.464) outperform other models, while for Mpox ($K=10$), BERTopic (.527) achieves the highest alignment, followed by BTM (.471). These results show that \textit{DoTA} captures meaningful differences in performance that are not reflected by traditional topic-based metrics.

Among the extended \textit{DoTA} variants, $DoTA_{\text{max}}$ first accounts for the strongest competing topic and reflects confidence in topic assignment under the hardest case. For example, LDA achieves the highest $DoTA_{\text{max}}$ in DoxyPEP at $K=5$ (.126), indicating clearer separation from its most competitive alternative, whereas TSCTM shows a much lower margin in COVID-19 at $K=10$ (-.142), suggesting greater assignment ambiguity. The negative $DoTA_{max}$ values observed in many settings indicate that the strongest competing topic is, on average, more similar than the assigned topic under the $DoTA_{max}$ formulation, reflecting assignment ambiguity. Thus, less negative values indicate a smaller competitive deficit rather than strong absolute assignment quality. Beyond the single strongest competitor, $DoTA_{\text{disc}}$ evaluates separation from all alternative topics on average. For instance, in the COVID-19 dataset ($K=10$), FASTopic achieves the highest $DoTA_{\text{disc}}$ (.111), indicating stronger overall separation despite lower absolute alignment. Similarly, in the DoxyPEP dataset ($K=10$), BTM attains the highest $DoTA_{\text{disc}}$ (.215), suggesting improved discriminative separation under increased topic granularity.

Overall, these results demonstrate that conventional topic-based metrics primarily evaluate topic structure and keyword distribution, whereas \textit{DoTA} metrics capture document–topic semantic alignment and assignment-related ambiguity. This distinction highlights a critical gap in existing evaluation frameworks and underscores the importance of incorporating assignment-aware metrics for short-text topic modeling, especially for public health-related social media posts. By providing complementary evaluation signals, \textit{DoTA} enables a more comprehensive and practically meaningful assessment of topic model performance across datasets and modeling paradigms.

\subsection{Sensitivity Analysis over Number of Topics for \textit{DoTA} Metrics}
\label{sentivity_study}

\begin{figure}[t]
\centering
\includegraphics[width=\linewidth]{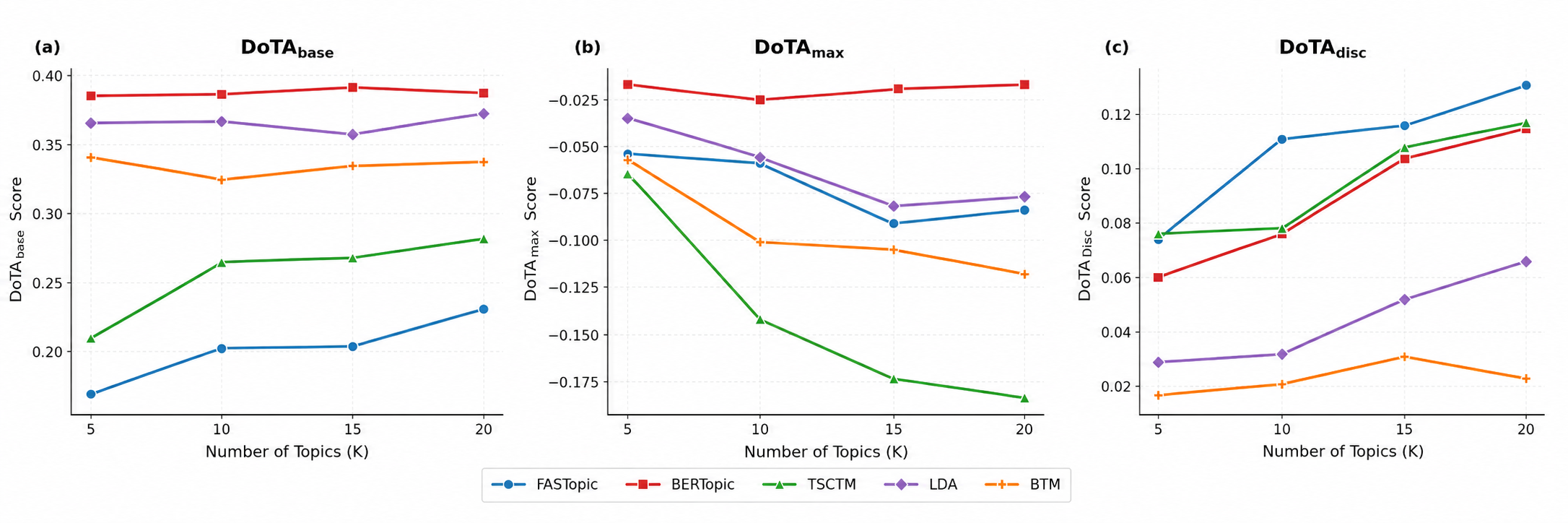}
\caption{Sensitivity of DoTA metrics across different topic numbers based on the COVID-19 dataset.}
\label{fig:sensitivity}
\end{figure}

We further examined the sensitivity of the \textit{DoTA} metrics to the number of topics $K$ on the COVID-19 dataset (Figure~\ref{fig:sensitivity}). The relative ranking of models remains largely stable across $K \in \{5, 10, 15, 20\}$, indicating robustness to topic granularity. BERTopic consistently achieves the highest $DoTA_{\text{base}}$ and $DoTA_{\text{max}}$, while FASTopic and TSCTM dominate $DoTA_{\text{disc}}$, and LDA remains competitive in $DoTA_{\text{max}}$. 



Although absolute values vary slightly with $K$, these changes do not alter the overall ordering of the topic models. Notably, $DoTA_{\text{base}}$ is relatively stable, $DoTA_{\text{max}}$ tends to decrease, reflecting stronger competition among topics as $K$ grows, and $DoTA_{\text{disc}}$ generally increases, indicating improved discriminative separation.

\subsection{Metrics Correlation Analysis}

\begin{figure}[t]
\centering
\includegraphics[width=\linewidth]{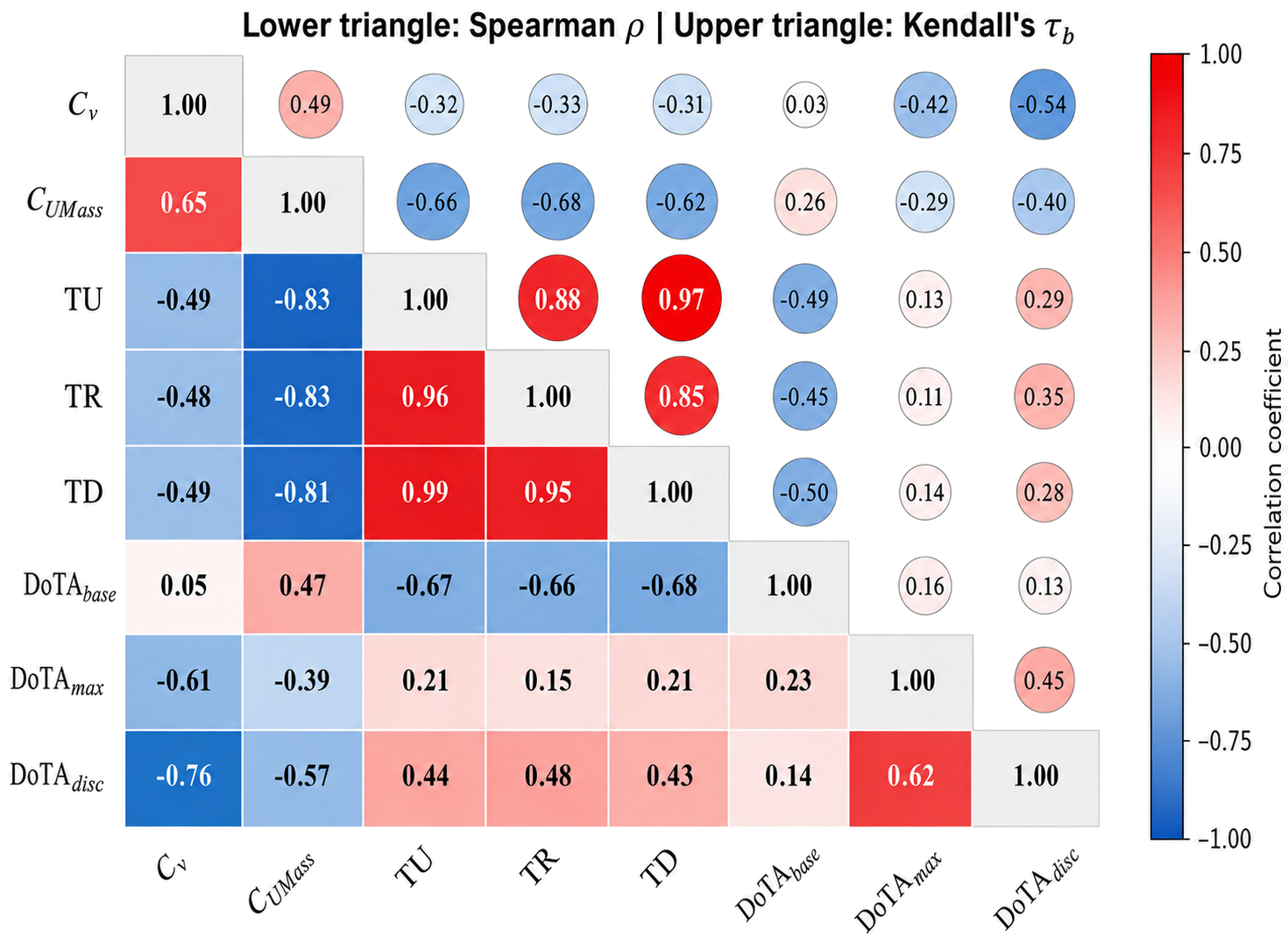}
\caption{Spearman's $\rho$ and Kendall's $\tau_b$ correlations between metrics. The lower triangle shows Spearman's $\rho$ as filled squares, and the upper triangle shows Kendall's $\tau_b$ as circles, with circle size indicating absolute correlation magnitude.}
\label{fig:spearman_kendall}
\end{figure}

Figure~\ref{fig:spearman_kendall} presents the Spearman's $\rho$ and Kendall’s $\tau_b$ correlations among topic-based and \textit{DoTA} metrics. These correlation coefficients were chosen because they capture monotonic relationships and are robust to nonlinear metric distributions. Both correlation measures exhibit highly consistent patterns, indicating robustness of the observed relationships.

Conventional topic metrics form a strongly correlated cluster (e.g., $TU$, $TR$, and $TD$). For both correlation measures, $TU$, $TR$, and $TD$ show very strong positive correlations (e.g., Spearman $>0.95$, Kendall $>0.85$), while $C_v$ and $C_{\text{UMass}}$ are also strongly correlated. These results suggest that conventional metrics largely capture similar aspects of topic structure, such as keyword overlap and coherence. In contrast, \textit{DoTA} metrics show weak or negative correlations with these topic-level metrics, indicating that document–topic alignment provides a complementary evaluation perspective. Notably, $DoTA_{\text{disc}}$ and $DoTA_{\text{max}}$ exhibit distinct correlation patterns, reinforcing their complementary roles. In addition, $DoTA_{\text{base}}$ exhibits weak or negative correlations with most topic-based metrics. For example, it shows near-zero correlation with $C_v$ (Spearman: $0.05$, Kendall: $0.03$) and strong negative correlations with $TU$, $TR$, and $TD$ (Spearman $\approx -0.67$, Kendall $\approx -0.45$ to $-0.50$). This indicates a clear divergence between topic-based quality and document-level semantic alignment. Moreover, the discriminative \textit{DoTA} variants behave differently. $DoTA_{\text{disc}}$ shows moderate positive correlations with $TU$, $TR$, and $TD$ (Spearman $\approx 0.43$–$0.48$, Kendall $\approx 0.28$–$0.35$), suggesting partial alignment with topic distinctiveness. In contrast, $DoTA_{\text{max}}$ exhibits weaker correlations, reflecting its focus on the most competitive alternative topic rather than overall topic structures. Notably, positive correlations within the \textit{DoTA} family (e.g., $DoTA_{\text{disc}}$ vs. $DoTA_{\text{max}}$: Spearman $0.62$, Kendall $0.45$) suggest internal consistency while still capturing distinct evaluation dimensions.

In summary, these results demonstrate that \textit{DoTA} metrics provide a complementary evaluation perspective that is not fully captured by traditional topic-based metrics, highlighting the importance of incorporating document-level alignment into topic model evaluation.



\subsection{Consistency with Human Ratings}

We further conducted an initial human validation on Cloud Research\footnote{\url{https://www.cloudresearch.com/}}. Native English-speaking raters evaluated 100 randomly sampled document–topic pairs (20 per model) on a 1–5 Likert scale. For $DoTA_{\text{max}}$ and $DoTA_{\text{disc}}$, participants also rated the most competitive alternative topic and a randomly sampled non-assigned topic, respectively. Each item was rated by five raters. Samples covered all three datasets and ratings were averaged across the five raters. Inter-rater agreement was assessed using unweighted Cohen’s $\kappa$. Table~\ref{tab:human_corr} reports the correlations between \textit{DoTA} and human ratings, with strong inter-rater agreement ($\kappa=0.82$).

\begin{table}[t]
\centering
\small
\caption{Correlations between DoTA metrics and human ratings (Cohen's $\kappa$ = 0.82)}
\label{tab:human_corr}
\begin{tabular}{lcc}
\toprule
\textbf{Metric} & \textbf{Spearman’s $\rho$} & \textbf{Kendall’s $\tau_b$} \\
\midrule
$DoTA_{\text{base}}$ & .82 & .73 \\
$DoTA_{\text{max}}$  & .63 & .52 \\
$DoTA_{\text{disc}}$ & .67 & .53 \\
\bottomrule
\end{tabular}
\end{table}

Based on Table~\ref{tab:human_corr}, all \textit{DoTA} metrics are positively correlated with human judgments, providing initial evidence of their consistency with perceived document--topic alignment. Among the three variants, $DoTA_{\text{base}}$ shows the strongest correlation (Spearman’s $\rho = 0.82$, Kendall’s $\tau_b = 0.73$), indicating the closest correspondence with human judgments of absolute semantic alignment. 

$DoTA_{\text{max}}$ and $DoTA_{\text{disc}}$ also show substantial correlations (Spearman’s $\rho = 0.63$ and $0.67$; Kendall’s $\tau_b = 0.52$ and $0.53$, respectively), suggesting that they capture human-perceived aspects of inter-topic competition and discrimination. Their lower correlations relative to $DoTA_{\text{base}}$ are consistent with their focus on more nuanced assignment relationships.

Overall, these initial findings suggest that \textit{DoTA}, particularly $DoTA_{\text{base}}$, aligns meaningfully with human judgments, while $DoTA_{\text{max}}$ and $DoTA_{\text{disc}}$ provide complementary signals for evaluating competing topic assignments.


\section{Conclusion}

This study proposed \textit{DoTA} metrics for evaluating topic modeling on social media short texts. By directly measuring how well an assigned topic semantically represents a document, \textit{DoTA} complements conventional topic-based metrics that focus only on topic coherence and distinctiveness. We introduced three variants, $DoTA_{\text{base}}$, $DoTA_{\text{max}}$, and $DoTA_{\text{disc}}$, to capture absolute alignment, margin-based separation from the strongest competing topic, and broader discriminative separation from all alternative topics.

Experiments on three public health-related datasets across five topic models showed that \textit{DoTA} provides informative and complementary evaluation signals beyond conventional metrics. Correlation analyses further showed that \textit{DoTA} captures information not fully reflected by topic-based measures, while human evaluation showed meaningful positive correlations with all \textit{DoTA} variants, especially $DoTA_{\text{base}}$.

Results demonstrate the importance of incorporating document-level semantic alignment into topic model evaluation. A limitation is that \textit{DoTA} remains conditional on the chosen embedding space, which may interact with model-internal representations. Future work should test alternative, decoupled encoders and broader domains.

\printbibliography

\end{document}